\documentclass[runningheads]{llncs}
\usepackage{graphicx}
\usepackage{amsmath}
\usepackage{amssymb}
\usepackage{floatrow} 
\usepackage{multirow}
\usepackage{mathtools}

\DeclareMathOperator*{\argmax}{argmax}
\begin{document}
\title{\large Is Aligning Embedding Spaces a Challenging Task?
 A Study on Heterogeneous Embedding Alignment Methods}
\titlerunning{An Analysis of Aligning Embedding Spaces methods}
%
\author{Russa Biswas\inst{1,2} \and Mehwish Alam\inst{1,2} \and
Harald Sack\inst{1,2}}
\authorrunning{Russa Biswas et al.} 
%
%
\institute{FIZ Karlsruhe -- Leibniz Institute for Information Infrastructure, Germany \\
\and
Karlsruhe Institute of Technology, Institute AIFB, Germany \\
\email{\{firstname.lastname\}@fiz-karlsruhe.de}}

\maketitle              
\begin{abstract}
Representation Learning of words and Knowledge Graphs (KG) into low dimensional vector spaces along with its applications to many real-world scenarios have recently gained momentum.
In order to make use of multiple KG embeddings for knowledge-driven applications such as question answering, named entity disambiguation, knowledge graph completion, etc., alignment of different KG embedding spaces is necessary.
In addition to multilinguality and domain-specific information, different KGs pose the problem of structural differences making the alignment of the KG embeddings more challenging.
This paper provides a theoretical analysis and comparison of the state-of-the-art alignment methods between two embedding spaces representing entity-entity and entity-word. This paper also aims at assessing the capability and short-comings of the existing alignment methods on pretext of different applications.

\end{abstract}

 \section{Introduction}
 
In recent years, there has been a rapid growth in the studies related to Representation Learning (RL), i.e., learning representations of input data by encoding different variations of the features. It plays a key role in performing machine learning tasks~\cite{bengio2013representation}.
The distributed representation of text in the form of word and document vectors~\cite{DBLP:journals/corr/abs-1301-3781} as well as of the entities and relations of a Knowledge Graph (KG) in the form of their corresponding vectors~\cite{DBLP:conf/nips/BordesUGWY13,DBLP:conf/aaai/LinLSLZ15} have evolved as the prime elements for various Natural Language Processing (NLP) tasks such as Entity Linking~\cite{DBLP:conf/esws/MorenoBBDLRTG17}, Named Entity Recognition and Disambiguation~\cite{yamada2016joint}, etc. 
Word embeddings are a low dimensional vector representation of words capable of capturing the context of a word in a document, semantic and syntactic similarity as well as its relation with other words. Similarly, KG embeddings are a low dimensional vector representation of entities and relations from a KG preserving its (local) inherent structure and capturing the semantic similarity between the entities. Therefore, each embedding space exhibits different characteristics based on the semantic differences in the source of information provided as input. 

It has been first observed in~\cite{DBLP:journals/corr/MikolovLS13} that continuous word embeddings exhibit similar structures across languages, even for the distant ones such as English and Vietnamese. The similarity has been exploited by learning a linear mapping from the source to a target embedding space. A parallel vocabulary with five thousand words as anchor points has been used to learn the mapping and evaluated against word translation. In~\cite{DBLP:conf/eacl/FaruquiD14,DBLP:conf/naacl/XingWLL15,DBLP:conf/emnlp/ArtetxeLA16}, the authors have attempted to improve cross lingual word embeddings based on bilingual word lexicons. Recent approaches aim to learn unsupervised alignment of monolingual word embedding spaces into a unified vector space
without using any parallel corpora~\cite{DBLP:conf/coling/CaoZZM16,conneau2017word}.

On the other hand, alignment of embedding spaces generated from heterogeneous input sources such as multiple KGs, KG and text, etc. have been well studied in \cite{hao2016joint,chen2017multilingual,sun2017cross,zhu2017iterative}. The alignment of the embedding spaces of two KGs refers to align the entities representing the same semantic concepts. 
In case of different KGs, the alignment would lead to links between multiple KGs. The key challenges encountered while aligning KG embeddings are as follows: {\bf (i)} Each KG is constructed differently, i.e., variations in the source of information or different ways of structuring the information. For instance, DBpedia
is constructed by automatically processing Wikipedia infoboxes, whereas Wikidata
is assembled based on collaborative efforts from its user base. {\bf (ii)} KGs often are multilingual, even though links between the same entities in different languages are already existing, but these links are still far from being complete.
{\bf (iii)} Multilingual KGs may contain complementary or even contradicting information about the same entity in different languages. 
{\bf (iv)} Multiple KGs of the same domain have a considerable overlap in their entities and relations and only some of the equivalent entity links exist between them. 

The context of an entity mention in textual resources is different from the contextual information of the same entity provided in the KG (i.e., through its surrounding nodes and edges). Textual context provides an insight into the pretext of where the entity has been mentioned in a certain document, whereas KG provides the structured information of the entity and its relation with other entities. Hence, joint training of word and entities in the same vector space yields effective results in NLP applications such as Named Entity Recognition and Disambiguation~\cite{yamada2016joint}.

This work focuses on the analysis of existing methods for aligning embedding spaces generated from heterogeneous input sources. The contributions include: 
\begin{itemize}
    \item A detailed description of the current state-of-the-art methods for aligning embedding spaces, their advantages and the challenges.
    \item The gaps in the existing research are discussed with an indication to directions of future work. 
\end{itemize}

To the best of our knowledge, this is the first study on the discussion and the comparison of the different alignment methods of embedding spaces. In this paper, the terms embedding space and vector space have been used interchangeably. The rest of the paper is organized as follows: Sect.~\ref{sec:problemformulation} presents the problem formulation. In Sect.~\ref{sec:alignment}, a detailed analysis of alignment methods for embedding spaces is provided, followed by the discussion of the results of the alignment in Sect.~\ref{sec:results}. Future work and conclusion is described in Sect.~\ref{sec:discussion_futurework}.
\section{Problem Formulation}
\label{sec:problemformulation}

In this study, the alignment models are categorized into two categories based on the type of the aligned spaces: {\bf(i)} Entity - Entity, and {\bf(ii)} Entity - Word

\begin{itemize}
    \item Entity - Entity: Independent embedding spaces generated from two KGs are aligned to form a common vector space.
    \item Entity - Word: Word and entity embeddings are learned jointly in the same vector space.

\end{itemize}
Since the embedding spaces are diverse, the challenges in the alignment task which have been analysed in this study are as follows:
\begin{itemize}
\item \textbf{RQ1} -- \textit{How to encode the structural differences of different resources (e.g., text and KG, different KGs such as DBpedia and Wikidata) in the aligned entity space?} 
\item \textbf{RQ2} -- \textit{How the heterogeneity of different KGs, in case of entity - entity embedding space alignment, are captured and represented in the resultant vector space?} 
\item \textbf{RQ3} -- \textit{How to combine the equivalent relations and its effect in the aligned embedding space generated from different KGs?}
\end{itemize}

\section{Alignment of Embedding Spaces}
\label{sec:alignment}
This section discusses the alignment methods for both the aforementioned categories of the algorithms proposed so far for embedding alignment, i.e., entity-entity and entity-word, followed by a discussion on the drawbacks based on the research questions.
\subsection{Entity - Entity Embedding Alignment}
\label{sec:entity-entity}

\textbf{JE}~\cite{hao2016joint} jointly learns embeddings of multiple KGs in a unified vector space with a focus on the entity alignment task in KGs. A set of already aligned entities between the KGs is used as a seed to learn embeddings through TransE~\cite{DBLP:conf/nips/BordesUGWY13}. The loss function is optimized by stochastic gradient descent and a projection matrix is introduced which serves as the transformation of different KG vector spaces. Considering two KGs, $G_{1}$ and $G_{2}$ the embeddings are learned by minimizing the margin-based objective function over the training set.
\begin{equation}
\begin{aligned}[2]
    L =   \sum_{(h,r,t) \in S}\sum_{(h',r,t') \in S'_{(h,r,t)}}\left\{\left[\gamma + d(h+r,t) - d(h'+r,t')\right]_+ 
+ \lambda_1 \sum_{y\in{\left\{ h,h',r,t,t'\right\} }}  | ||y_2|| -1 |\right\}  \\
+ \lambda_2 \sum_{(e_i,e_i')\in A} || e_i - e_i' ||_2,
\label{eq:1}
\end{aligned} 
\end{equation}
where $(h,r,t)$ denotes a triple, \([x]_+\) denotes the positive part, \(\gamma > 0\) is a margin-hyper-parameter, \(\lambda_1, \lambda_2\) are ratio hyper-parameters, $A$ is the selected seed alignment whose entities are represented by \(e_i\) in $G_1$ and \(e_i'\) in $G_2$ and 
\begin{equation}
    S'_{(h,r,t)} =  \left\{(h',r,t)|h' \in E\right\} \cup \left\{(h,r,t')|t' \in E \right.\}
\end{equation}
\(\lambda_1\) in Eq.\eqref{eq:1} plays the soft constraint of the entities and relations by trivially minimizing the loss function which is achieved by increasing the embedding norms. \(\lambda_2\) in Eq.\eqref{eq:1} helps in learning the alignment between the KGs. The transformation of the different KG vector spaces is done by a projection transformation matrix \(M_d\) and second part (\(\lambda_2\)) of the Eq.\eqref{eq:1} is modified to 
\[ \lambda_2 \sum_{(e_i,e_i')\in A} ||M_de_i - e_i' ||_2 . \]
The algorithm has been evaluated over two datasets: {\bf(i)} FB15K, a subset of Freebase KG~\cite{bollacker2008freebase} which is split into two parts where each split is considered as one KG. {\bf(ii)} DB-FB dataset consists of a subset of triples from DBpedia and Freebase. The results depict that higher the number of aligned seeds the better the performance of the model. The model performs well for the DB-FB dataset, however, it is difficult to capture the accuracy of the model for the whole DBpedia and Freebase due to sparsity issues.

On the other hand, \textbf{MTransE}~\cite{chen2017multilingual} is a translation based multilingual KG embedding model. As a first step it encodes entities and relations of each language in a separate vector space using TransE, followed by transition of each vector to its corresponding cross lingual counterpart in the other space. The loss function is the weighted sum of the knowledge model and the alignment model and is given by,
\begin{equation}
    J = S_K + \alpha S_A,
\end{equation}
where \(S_K\) is the knowledge model which is generated by using TransE for each language, \(\alpha\) is a hyper-parameter that weights the alignment model \(S_A\). 
The loss function measures the plausibility of all given triples and is given by,
\begin{equation}
    S_K = \sum_{L \in \left\{L_i, L_j\right\}}\sum_{(h,r,t) \in G_L} ||h + r - t ||,
\end{equation}
where $L$ represents the set of languages in a multilingual KG, $G$. The model is trained on partially aligned graphs, i.e., aligned triples for the cross lingual KGs. The loss function for the alignment model is given by,
\begin{equation}
    S_A = \sum_{(T,T') \in \delta(L_i,L_j)}S_a(T,T'),
\end{equation}
where the alignment score \(S_a(T,T')\) iterates through all pairs of aligned triples. Three different techniques have been used to align the vector spaces, i.e., distance-based axis calibration, translation vectors and linear transformation. The distance-based axis calibration penalizes the alignment based on the distances of cross-lingual counterparts such that multilingual expressions of the same entity are closer together after alignment using,
\begin{equation}
    S_{a_1} = ||h - h'|| + ||t - t'||.
\end{equation}
The coordinates of the same relation in multilingual KGs are converged using,
\begin{equation}
    S_{a_2} = ||h - h'|| + ||r - r'|| + ||t - t'||.
\end{equation}
The translation vector model consolidates alignment into graph structures and characterizes cross-lingual transitions as regular relational translations and is given by,
\begin{equation}
    S_{a_3} = ||h + v_{ij}^e - h' || + ||r + v_{ij}^r - r' || + ||t + v_{ij}^t - t' ||,
\end{equation}
where \(v_{ij}^e\) and \(v_{ij}^r\) are the entity-dedicated and relation-dedicated translation vectors between \(L_i\) and \(L_j\), such that \(e + v_{ij}^e \approx e'\) for embedding vectors $e$, $e'$ of the same entity in both languages. Similar translation holds for relations. Finally, the linear transformation model considers transitions as topological transformation of embedding spaces without assuming the similarity of spatial emergence and is defined as,
\begin{equation}
    S_{a_4} = ||M_{ij}^eh - h'|| + ||M_{ij}^et - t'||,
\end{equation}
where \(M_{ij}^e\) is a \(k \times k\) square matrix learned as a linear transformation on entity vectors from \(L_i\) to \(L_j\) and $k$ is the dimensionality of the embedding space. For relation, the transformation model is modified to,
\begin{equation}
    S_{a_5} = ||M_{ij}^eh - h'|| + ||M_{ij}^rr - r'|| + ||M_{ij}^et - t'||.
\end{equation}
The model has been evaluated against WK31-15K, a dataset containing entities from English, French and German versions. The experiments imply that the variants using linear transformation work better in cross lingual entity matching. The key difference between JE and MTransE is that JE uses aligned entities as seeds whereas MTransE uses aligned triples as seeds.

\textbf{Joint Attribute Preserving Embedding (JAPE)}~\cite{sun2017cross} combines structure embedding and attribute embedding to align entities in different KGs. A set of aligned entities between the two KGs are considered as the seeds for this model. TransE is used for generating structure embedding whereas Skip-gram is used for attribute embedding. Each pair in the seed alignment share the same representation to serve as a bridge between \(G_1\) and \(G_2\) to build an overlay relationship graph. The objective function for the structure embedding is,
\begin{equation}
    O_{SE} = \sum_{tr \in T} \sum_{(tr' \in T'_{tr})}(f(tr) - \alpha f(tr')),
\end{equation}
where $T$ denotes the set of all positive triples, i.e., the existing triples and \(T'_{tr}\) denotes the associated negative triples for $tr$ generated by replacing either the head or the tail by random entity. \(\alpha\) is the ratio hyper-parameter that weights positive and negative triples in the range of [0,1]. For attribute embedding, given an aligned entity pair \((e_1, e_2)\), it is assumed that the attributes of \(e_1\) and \(e_2\) are highly correlated to each other. Skip-gram model is used to predict correlated attributes for a certain attribute minimizing the loss function,
\begin{equation}
    O_{AE} = - \sum_{(a,c) \in H} \omega_{a,c}.log p(c|a),
\end{equation}
where $H$ denotes the set of positive $(a,c)$ pairs, i.e., $C$ is actually correlated attribute of $a$, and the term $logp(c|a)$ denotes the probability. The joint attribute preserving embedding uses matrices of pairwise similarities between entities as supervised information and minimizes the following object function,
\begin{equation}
    O_S = ||E_{SE}^{(1)} - S^{(1,2)}E_{SE}^{(2}||_F^2 + \beta(||E_{SE}^{(1)} - S^{(1)}E_{SE}^{(1)}||_F^2 + ||E_{SE}^{(2)} - S^{(2)}E_{SE}^{(2)}||_F^2),
\end{equation}
where \(\beta\) is a hyper-parameter that balances similarities between KGs and their inner similarities. \(E_{SE} \in \mathbb{R}^{n_e \times d}\) denotes the matrix of entity vectors for one KG in S with each row as entity vector. \(S^{(1,2)}E_{SE}^{(2}\) calculates the latent vectors of entities in \(G_1\) by accumulating vectors of entities in \(G_1\) based on their similarities measured by using cosine similarity.
To preserve both the structure and attribute information, the following objective function is minimized,
\begin{equation}
    O_{joint} = O_{SE} + \delta O_S
\end{equation}
where \(\delta\) is the hyper-parameter weighting \(O_S\). 
The model is evaluated against the DBP15K dataset which is composed of Chinese, Japanese, French and English DBpedia versions. It is used for cross lingual entity alignment tasks from different languages to English. The attribute embedding provides additional information which leads to better performance of the entity alignment task as compared to the previous methods.

\textbf{Iterative TransE (ITransE)}~\cite{zhu2017iterative} first learns both entity and relation embeddings, based on aligned entity seeds using TransE followed by a mapping of the embeddings from different KGs. The pre-requisite of this model is that both  KGs must contain the same relations. The model introduces a translation model, a linear transformation model, and a parameter sharing model. For the translation model, given two aligned entities \(e_1 \in E_1\) and \(e_2 \in E_2\), it is assumed that there exists an alignment relation \(r^{(E_1 \rightarrow E_2)}\) such that \(e_1 + r^{(E_1 \rightarrow E_2)} \simeq e_2\). Therefore, the joint embedding is defined by,
\begin{equation}
    E(e_1,e_2) = || e_1 + r^{(E_1 \rightarrow E_2)} - e_2 ||,
\end{equation}
where \(E_1\) and \(E_2\) denote the set of entities from $G_1$ and $G_2$ respectively. In the linear transformation model, for the two given aligned entities as mentioned above, a transformation matrix \(M^{(E_1 \rightarrow E_2)}\) so that \(M^{(E_1 \rightarrow E_2)}e_1 \simeq e_2\). Hence the energy function is defined by,
\begin{equation}
    E(e_1,e_2) = ||M^{(E_1 \rightarrow E_2)}e_1 - e_2 ||.
\end{equation}
For both the aforementioned models, the score function is defined as the sum of the energy functions over alignment seeds and is given by,
\begin{equation}
    \tau_{T/L} = \sum_{(e_1,e_2) \in L} \alpha E(e_1,e_2),
\end{equation}
where \(\alpha\) is the weighted factor and L is the set of aligned seeds. On the other hand, in parameter sharing model, each aligned entity pair \(e_1, e_2\) is defined as \(e_1 \equiv e_2, (e_1,e_2) \in L\) and the distance is measured by,
\begin{equation}
    E(e_1,e_2) = ||e_1 - e_2||_{L1/L2}, \forall e_1 \in E_1, e_2 \in E_2.
\end{equation}
Since no regularisation is involved in this model, the scoring function is \(\tau = 0\). Hence, for each non-aligned entity \(e_1\) in one KG, the nearest non aligned entity \(\hat{e_2}\) from another KG is given by \(\hat{e_2} = argmin_{e_2}(E(e_1,e_2))\), which is the newly found entity. These newly discovered entity alignments are included in the initial set of alignment seeds \(L\) to update the joint embedding. There are two variations of this inclusion of the newly discovered seeds. If all the seeds are added to \(L\), it is referred to as \emph{hard alignment}. But if a reliability score is added to each of these newly discovered aligned entities while including them to the set \(L\), then it is called \emph{soft alignment}. The experiments on the entity alignment task have been carried out on four subsets of FB15K dataset containing overlapping triples. The parameter sharing model outperforms linear transformation and translation models. 

Another method has been proposed for cross lingual KG alignment using Graph Convolutional Network (GCN)~\cite{wang2018cross}. This method trains GCNs to embed entities of each language to a unified vector space using pre-aligned entities. Relation alignment across KGs is not required to train the model. Entity embeddings are learned from both structural and attribute information of the entities and alignments are discovered based on the distances between them in the vector space. In this approach, two GCN models process two KGs independently to generate the embeddings. Both the GCNs share the same weight matrices. The first layer of each GCN model transforms the input attribute feature vectors and two GCN-models generate attribute embeddings of the same dimensionality. Entity alignment are predicted based on distances between entities from two KGs in the GCN representation space. For entity \(e_i\) in $G_1$ and \(v_j\) in $G_2$, the distance measure is computed by,
\begin{equation}
    D(e_i,v_j) = \beta \frac{f(h_s(e_i),h_s(v_j))}{d_s} + (1 - \beta) \frac{f(h_a(e_i),h_a(v_j))}{d_a},
\end{equation}
where \(f(x,y) = || x - y ||_1\), \(h_s(\cdot)\) and \(h_a(\cdot)\) denote the structure embedding and attribute embeddings of an entity respectively. \(d_s\) and \(d_a\) denote the dimensions of structure and attribute embedding respectively, and \(\beta\) is the hyper-parameter that balances the importance of both kinds of embeddings. The GCN models are trained by minimizing the following margin-based ranking loss function on a set of pre-aligned entities $S$ as training data,
\begin{equation}
    L_s = \sum_{(e,v) \in S} \sum_{(e',v') \in S'_{(e,v)}} [f(h_s(e),h_s(v)) + \gamma_s - f(h_s(e'),h_s(v'))]_+  
\end{equation}
and
\begin{equation}
    L_a = \sum_{(e,v) \in S} \sum_{(e',v') \in S'_{(e,v)}} [f(h_a(e),h_a(v)) + \gamma_a - f(h_a(e'),h_a(v'))]_+  ,
\end{equation}
where \(S'_{(e,v)}\) denotes the set of negative triples and \(\gamma_s, \gamma_a > 0\) are margin hyper-parameters separating positive and negative entity alignments. The experiments of the alignment task have been performed on DBP15K dataset for the languages Chinese - English, Japanese - English, and French - English.

\textbf{Bootstrapping Entity Alignment}~\cite{sun2018bootstrapping} models the alignment as a classification problem which seeks to maximize alignment likelihood over all labeled and unlabeled entities based on KG embeddings. Let $X$ and $Y$ be the entity sets of \(G_1\) and \(G2\) respectively. It considers only one-to-one alignment, i.e., entities in Y are used to label entities in X and only one label is assigned to one entity in X. In order to ensure that the positive triples (aligned entities) have low scores and the negative triples have high scores, the following limit-based objective function is proposed,
\begin{equation}
    O_e = \sum_{(\tau \in \mathbb{T}^+)}[f(\tau) - \gamma_1]_+ + \mu_1\sum_{(\tau' \in \mathbb{T}^-)}[\gamma_2 - f(\tau')]_+,
\end{equation}
where \([\cdot]_+ = max(\cdot,0)\), \(\gamma_1,\gamma_2\) are two hyper-parameters, \(\mu_1 > 0\) is a balance hyper-parameter, and \(\mathbb{T}^+, \mathbb{T}^-\) denote the set of positive and negative triples respectively. Therefore, the drift of the entity embeddings in the unified space is reduced and common semantics of the two KGs are better captured. To leverage prior alignment, for bridging different KGs, aligned entities in their triples are swapped to calibrate the embeddings of \(G_1\) and \(G_2\) in the unified embedding space. However, there is often inadequate prior alignments which is overcome by bootstrapping. Obeying one-to-one alignment constraint and to label the alignment in the \textit{t}-th iteration, the optimization problem is solved as follows:
\begin{equation}
\begin{split}
     max \sum_{x \in X'} \sum_{y \in Y'_x} \pi(y|x;\theta^{(t)})\cdot \psi^{(t)}(x,y),\\
    s.t. \sum_{x' \in X'} \psi^{(t)}(x',y) \leq 1, 
    \sum_{y' \in Y'_x} \psi^{(t)}(x,y') \leq 1, \forall x,y,
\end{split}
\end{equation}
where \(\theta^{(t)}\) denotes the entity embeddings at the \textit{t}-th iteration, \(Y'_x\) denotes candidate labels of \(x\), and \(\psi^{(t)}(\cdot)\) is an indicator function. \(\psi^{(t)}(x,y) = 1\) if \(x\) is labeled as \(y\) at the \textit{t}-th iteration and 0 otherwise. The model is evaluated against DBP15K, and DWY100K which comprises of DBP-WD (DBpedia and Wikidata) and DBP-YG (DBpedia and YAGO). It outperforms the state-of-the-art methods, MTransE, ITransE, and JAPE.

\textbf{Attribute embedding based entity alignment}~\cite{trisedya2019entity} framework consists of three components namely predicate alignment, embedding learning, and entity alignment. The predicate alignment module merges two KGs into one KG by renaming the partially matched predicates with a similarity threshold of 0.95. Next, the embedding learning phase is comprised of structure embedding and attribute character embedding. TransE has been adapted for the structure embedding of the newly generated merged KG. The following objective function is minimized for the structure embedding,
\begin{equation}
    J_{SE} = \sum_{t_r \in T_r}\sum_{t'_r \in T'_r} max(0, [\gamma + \alpha(f(t_r) - f(t'_r))]),
\end{equation}
and
\begin{equation}
    \alpha = \frac{count(r)}{|T|},
\end{equation}
where \(T_r\), and \(T'_r\) are the set of valid and corrupted relation triples respectively, count(r) is the total number of occurrences of a relation \(r\), \(|T|\) is the total number of triples in the merged KG, and \(\alpha\) is the weight introduced which controls the embedding learning over the triples. To learn the attribute character embedding, the objective function is given by,
\begin{equation}
    J_{CE} = \sum_{t_a \in T_a}\sum_{t'_a \in T'_a} max(0, [\gamma + \alpha(f(t_a) - f(t'_a))]),
\end{equation}
\begin{equation}
    T_a = \left\{(h,r,a) \in G\right\}; f(t_a) = ||h + r -f_a(a)||,
\end{equation}
\begin{equation}
    T'_a = \left\{(h',r,a)|h' \in E\right\}\cup\left\{(h,r,a')|a' \in A \right\} ,
\end{equation}
where \(T_a\) and \(T'_a\) are the set of valid and corrupted attribute triples and \(f_a(\cdot)\) is a compositional function. SUM is one of the compositional functions which sums up all the character values of the attribute value. On the contrary, LSTM based function converts a sequence of characters into a vector. 
Finally, attribute character embedding \(h_{ce}\) is used to shift the structure embedding \(h_{se}\) into the same vector space by minimizing the following objective function,
\begin{equation}
    J_{SIM} = \sum_{h \in G_1 \cup G_2} [1 - cos(h_{se}, h_{ce})].
\end{equation}
The overall objective function is given by,
\begin{equation}
    J = J_{SE} + J_{CE} + J_{SIM}.
\end{equation}
The final module addresses the entity alignment using,
\begin{equation}
    z = \argmax\limits_{h_2 \in G_2} cos(h_!, h_2),
\end{equation}
where for an entity \(h_1 \in G_1\), the similarities between \(h_1\) and all entities \(h_2 \in G_2\) are computed. \(<h_1, h_{map}>\) is the expected pair of aligned entities. Also, for improving the character embeddings, attribute triple enrichment is performed by transitivity rule. The model is evaluated against datasets generated from multiple heterogeneous KGs, i.e., DBpedia, YAGO and Geonames which outperforms all the current state of the art methods.

\paragraph{\bf{Challenges.}}The challenges of the above mentioned models are: {\bf (i)} They are supervised and require a set of aligned entities or triples as seeds for training, meaning that, parallel data is necessary for these methods to work. {\bf (ii)} Some of the models such as ITransE require the relations to be aligned between the KGs. However, in case of heterogeneous KGs such as DBpedia and Wikidata which consist of different sets of relations, it is a challenging task to have pre-aligned set of relations. {\bf (iii)} 
Equivalent links (such as {\tt owl:equivalentProperty}) between relations exist across KGs and none of the aforementioned methods exploit these links for better alignment. {\bf (iv)} The methods considering attribute information for the alignment fail to exploit the type of the attributes such as text literals, numeric literals, etc. {\bf (v)} The methods lack proper mechanisms to handle multi-valued object and attribute relations. 

\subsection{Entity - Word Embedding Alignment}
\label{sec:entity-word}
\textbf{CONV-augmented model}~\cite{toutanova2015representing} learns embeddings of words and entities into a single vector space using Convolutional Neural Network (CNN) with DistMult~\cite{yang2014embedding} for KG completion. CNN is trained on lexicalised dependency paths which are generated from the textual data treated as a sequence of words. The sentences are then annotated with the entities from the KG. The loss function is motivated by link prediction task and is given by,
\begin{equation}
    L(T;\Theta) = - \sum_{(e_s,r,e_o) \in T} logp(e_o|e_s,r;\Theta) -  \sum_{(e_s,r,e_o) \in T} logp(e_s|e_o,r;\Theta),
\end{equation}
where $T$ is the set of triples and \(e_s,e_o\) denote the subject and object entities. Let $T_{KB}$ and $T_{text}$ represent the set of knowledge base triples and textual relation triples respectively. Therefore the final loss function is defined as:
\begin{equation}
    L(T_{KB}; \Theta) + \tau L (T_{text}; \Theta) + \lambda||\Theta||^2,
\end{equation}
where \(\lambda\) is the regularization parameter and \(\tau\) is the weighing factor of the textual relations. Therefore, CNN model learns the latent representation of words and entities based on the annotated dataset generated from FB15K-237 and ClueWeb12
~\cite{} ~\cite{gabrilovich2013facc1}.

Similarly, \cite{han2016joint} uses TransE and CNN for joint learning of the entities and words in a unified vector space. TransE is used to learn the representation of the entities and the relations from the KG and the loss function w.r.t. this model is given by,
\begin{equation}
    L(G) = \sum_{(h,r,t) \in T}\sum_{(h',r',t') \in T'} [\gamma + f_r(h,t) - f_r'(h',t')]_+,
\end{equation}
where \([x]_+\) indicates keeping the positive part
and \(\gamma > 0\) is a margin, and $T'$ is the set of incorrect triples. To encode the text, given a sentence containing $(h,t)$ with a relation $r$, the model takes word embeddings \(s = \left\{X_1,X2,...,X_n\right\}\) of the sentence $s$ as input, and after passing through CNN, outputs the embedding of the textual relation $r_s$. The model further minimizes the loss between $r$ and $r_s$ which is formalized by,
\begin{equation}
    f_r(s) = ||r_s - r||_2,
\end{equation}
The loss function over all the sentences in the text dataset $D$ is given by,
\begin{equation}
    L(D) = \sum_{s \in D}\sum_{r' \neq r}[\gamma + f_r(s) - f_r'(s)]_+,
\end{equation}
For a word \(x_i\) in the given sentence, its input embedding \(X_i\) is composed of its textual word embedding \(w_i\) and its position embedding \(p_i\). Word position embedding of a word is generated from the position of that word in a sentence. Textual word embeddings are generated from pre-training of the text using skip-gram model. The CNN model comprises of an input layer, a convolutional layer and a pooling layer. The model
learns the embedding using CNN on Freebase and New York times articles. 


\cite{yamada2016joint} comprises of three models namely skip-gram model for word similarity, KB model, and anchor context similarity to jointly map the entities and words into a continuous vector space. The training of the overall model focuses on maximizing the following objective function such that the resulting matrix $V$ to embed the entities and words,
\begin{equation}
    L = L_w + L_e + L_a.
\end{equation}
The model extends the skip-gram model by using {\bf (i)} KB graph model which learns the relations between the entities using the link structure, and {\bf (ii)} anchor context model which aligns similar words and entities closer in the vector space by leveraging the KG anchors and context words.
Given a sequence of $T$ words $w_1,w_2,...w_T$, the model aims to maximize the following objective function,
\begin{equation}
    L_w = \sum_{t=1}^T \sum_{-c \leq j \leq c , j \neq 0 } log P (w_{t+j}|w_t),
\end{equation}
where $c$ is the size of the context window, and $w_t, w_{t+j}$ denote the target and the context word respectively. In the KB model, entities with similar incoming links are placed closer to each other in the vector space and is formalised as,
\begin{equation}
    L_e = \sum_{e_i \in E} \sum_{e_o \in C_{e_i}, e_i \neq e_o} log P(e_o|e_i).
\end{equation}
It is used to predict the incoming links $C_e$ given an entity $e$. Finally, in the anchor text model, the model is trained to predict the context words of an entity pointed to by the target anchor. The objective function is as follows,
\begin{equation}
    L_a = \sum_{(e_i, Q) \in A}\sum_{w_o \in Q}log P(w_o|e_i),
\end{equation}
where A denotes a set of anchors in the KB, each of which contains a pair of a referent entity $e_i$ and a set of its context words $Q$. Wikipedia is used as a KG to train the model and evaluated against CoNLL and TAC2010 datasets for named entity disambiguation task. However, this approach suffers from ambiguity, i.e., same words or phrases can refer to different entities and vice-versa.

To address this problem, \textbf{Multi-Prototype Mention Embedding (MPME)} model~\cite{cao2017bridge} has been proposed which learns multiple sense embeddings for each entity mention by jointly modeling words from textual contexts and entities derived from a KG. Furthermore, a language model is used to disambiguate each mention to a sense. The goal of training MPME is to maximize the function below, and iteratively update three types of embeddings,
\begin{equation}
    L = L_w + L_e + L_m,
\end{equation}
where $L_w$ denotes the word embeddings, $L_e$ denotes the entity embeddings, and $L_m$ denotes the word-mention representations. For entity representation, the skip-gram model is extended to a network by maximizing the log probability of being a neighbour entity and is formalized by,
\begin{equation}
    L_e = \sum_{e_j \in E}log P(N(e_j)|e_j).
\end{equation}
This is because the neighboring entities exhibit similar role as context words in the skip-gram model. Next, given an anchor $< m_l, e_j >$  and its context words $C(m_l)$, we combine mention sense embeddings with its context word embeddings to predict the reference entity by extending CBOW model. The objective function is as follows:
\begin{equation}
    L_m = \sum_{(<m_l,e_j>) \in A} log P(e_j|C(m_l),s_j^*),
\end{equation}
where \(s_j^* = g(<m_l,e_j>)\). If two mentions refer to similar entities and share similar contexts, they tend to be close in a vector space. On the other hand, given the annotated corpus $D'$, word $w_i$ or a mention sense $s_j^*$ is used to predict the context words by maximizing the following objective function,
\begin{equation}
    L_w = \sum_{w_i,m_l \in D'} log P(C(w_i)|w_i) + log P(C(m_l)|s_j^*),
\end{equation}
where \(s_j^* = g(<m_l,e_j>)\) is obtained from anchors of Wikipedia articles. The model is trained on Wikipedia and was evaluated for entity linking, entity relatedness and word analogy tasks. 

\cite{cao2018joint} proposes a method in which joint representation learning of cross-lingual words and entities via distant supervision is performed using multi-lingual KGs. It is based on assumption that the more words and or entities two contexts share the more similar they are. The framework comprises of two steps: {\bf (i)} Cross Lingual Supervision Data Generation which builds a bilingual network and generates comparable sentences based on cross-lingual links. Two regularizers are used to align the cross-lingual words and entities. The framework is built on the assumption that the more number of words or entities two context share, the more similar they are. The cross-lingual entity regularizer is given by,
\begin{equation}
    L_e = \sum_{e_i^y \in \left\{G^{en-zh}\right\}} log P(C'(e_i^y)|e_i^y),
\end{equation}
where $C'(e_i^y)$ denotes cross-lingual contexts neighbor entities in different languages that linked to $e_i^y$. The second regularizer is the cross lingual sentence regularizer. Comparable sentences provide cross-lingual co-occurrence of words which is used to learn similar embeddings for the words that frequently co-occur by minimizing the Euclidean distance as follows,
\begin{equation}
    L_s = \sum_{<s_k^en,s_k'^zh> \in S^{en-zh}} ||s_k^en - s_k'^zh||^2,
\end{equation}
where $s_k^en, s_k'^zh$ are sentence embeddings. {\bf (ii)} Joint Representation Learning, which learns the cross-lingual entities and words into a vector space. This model is also trained on Wikipedia and has been evaluated for natural language translation.

\paragraph{\bf{Challenges.}}The challenges of the entity -  word alignment models are as follows: {\bf (i)} DNN based models
suffer from huge vocabulary size of the words and entities as well as expensive training. {\bf (ii)} The models are supervised and require manual annotation of the text with the entities from the KG. {\bf (iii)} Most of the models use Wikipedia as the KG and generate an entity graph based on the Wikipedia hyperlinks and propose different ways of identifying the relevant context information. As a result only context words are projected in the vector space. {\bf (iii)} Models which consider external information from the KGs such as Freebase use only a fraction of the triples available for all the entities. Hence, all the information encoded in the KG for an entity is not exploited.

\section{Results and Discussion}
\label{sec:results}
This section summarizes the results of the methods discussed in the current survey as provided in their respective studies. These methods have been evaluated over the following tasks: entity alignment, KG completion, Named entity disambiguation (NED). 


\noindent\textbf{Datasets}: Entity-entity alignment models are evaluated on real KGs such as DBpedia (DBP), LinkedGeoData(LGD), Geonames(GEO) and YAGO datasets. As discussed in \cite{trisedya2019entity}, three models JAPE, MTransE, and entity alignment using attribute embeddings are tested against 
DBP-LGD, DBP-GEO, DBP-YAGO which contain aligned entities\footnote{\url{http://downloads.dbpedia.org/2016-10/links/}}. DBP-YAGO dataset contains 15,000 aligned entities whereas both DBP-LGD and DBP-GEO contain 10,000 aligned entities each.  On the other hand, cross lingual entity alignment task has also been performed using the models MTransE, ITransE, JAPE, and BootEA on the DBP15K dataset. The statistics of these datasets are given in Table~\ref{tab:statistics_dataset}.
\begin{table}[t!]
\begin{center}
\begin{tabular}{ |c|c|c|c|c| } 
\hline
\multicolumn{2}{|c|}{Dataset} & Entities & \parbox{2cm}{Attribute Triples} & \parbox{2cm}{Relationship Triples}\\
\hline
\multirow{2}{6em}{DBP-LGD} & LGD & 24,309 & 90,054 & 10,084 \\ 
& DBP & 22,748 & 166,008 & 19,594 \\ \hline
\multirow{2}{6em}{DBP-GEO} & GEO & 21,794 & 98,790 & 17,410 \\ 
& DBP & 22,748 & 166,008 & 19,594 \\  \hline 

\multirow{2}{6em}{DBP-YAGO} & YAGO & 30,628 & 173,309 & 38,451 \\ 
& DBP & 33,627 & 184,672 & 36,906 \\  \hline

\multirow{2}{6em}{DBP15K (ZH-EN)} & Chinese & 66,469 & 379,684 & 153,929 \\ 
& English & 98,125 & 567,755 & 237,674 \\  \hline
\multirow{2}{6em}{DBP15K (JA-EN)} & Japanese & 65,744 & 354,619 & 164,373 \\ 
& English & 95,680 & 497,230 & 233,319 \\  \hline
\multirow{2}{6em}{DBP15K (FR-EN)} & French & 66,858 & 582,665 & 192,191 \\ 
& English & 105,889 & 576,543 & 278,590 \\
\hline
\end{tabular}
\caption{Dataset Statistics}

\label{tab:statistics_dataset}
\end{center}
\vspace{-0.6cm}
\end{table}
\begin{table}
    \begin{tabular}{llllllllll}
       \hline
       Models & \multicolumn{3}{c}{DBP-LGD} &             \multicolumn{3}{c}{DBP-GEO} & \multicolumn{3}{c}{DBP-YAGO} \\
   &Hits@1 & Hits@10 & MR          & Hits@1 & Hits@10 & MR & Hits@1 & Hits@10 & MR \\
    \hline
    MTransE & 33.29 & 34.32 & 10194 & 33.34 & 33.98 & 10240 & 33.46 & 34.32 & 7105 \\
    JAPE & 33.33 & 33.35 & 5104 & 33.35 & 33.75 & 5088 & 33.35 & 33.37 & 5296 \\
    $N-gram_{attr}$ & 84.27 & 91.85 & 53 & 87.61 & 92.15 & 80 & 89.69 & 95.83 & 23 \\
    \hline
    \hline
     Models & \multicolumn{3}{c}{DBP(ZH-EN)} &             \multicolumn{3}{c}{DBP(JA-EN)} & \multicolumn{3}{c}{DBP(FR-EN)} \\
   &Hits@1 & Hits@10 & MRR          & Hits@1 & Hits@10 & MRR & Hits@1 & Hits@10 & MRR \\
    \hline
    MTransE & 30.83 & 61.41 & 0.364 & 27.86 & 57.45 & 0.349 & 24.41 & 55.55 & 0.335 \\
    ITransE & 40.59 & 73.47 & 0.516 & 36.69 & 69.26 & 0.474 & 33.30 & 68.54 & 0.451 \\
    JAPE & 41.18 & 74.46 & 0.49 & 36.25 & 68.50 & 0.476 & 32.39 & 66.68 & 0.43 \\
    BootEA & 62.94 & 84.75 & 0.703 & 62.23 & 85.39 & 0.701 & 65.30 & 87.44 & 0.731\\
    GCN & 41.25 & 74.38 & - & 39.91 & 74.46 & - & 37.29 & 74.49 & -\\
    \hline
    \end{tabular}
    \caption{Entity Alignment Task}
    \label{tab:entity_alignment}
    \vspace{-0.2cm}
\end{table}
\vspace{-0.6cm}

Intuitively, it is easier to align entities with more content. The same has been reflected in the results presented in Table~\ref{tab:entity_alignment}. The attribute information plays a vital role in the alignment task as it captures more semantics of the entities in a KG. Hence, $N-gram_{attr}$ and BootEA perform better than the other models. Also, it has been noticed that more the number of aligned seeds the better the performance of the model. The main advantage of the GCN model is that huge number of aligned seeds are not required for training. The values for MRR are not provided in the original paper, hence it is not presented in Table~\ref{tab:entity_alignment}.

On the other hand, the presented entity-word embedding space alignment models are evaluated for different tasks such as entity relatedness, entity linking, NED, KG completion, and word translation. Since the models are trained with a focus to perform a certain task, no two aforementioned models use the same dataset. However, CONV-augmented model~\cite{toutanova2015representing} and joint model of TransE with CNN~\cite{han2016joint} focus on the link prediction task. But CONV-augmented model uses FB15K-237 dataset whereas the later uses FB15K dataset. It is noticed that both the models work better when textual data is included. For NED, \cite{yamada2016joint} uses CoNLL and TAC2010 datasets and the results outperforms the state-of-the-art NED models. MPME~\cite{cao2017bridge} is evaluated for entity relatedness and word analogical reasoning. Lastly, the model \cite{cao2018joint} is compared with MPME for entity relatedness task for the multilingual datasets and it works slightly better than MPME. 

\section{Conclusion}
\label{sec:discussion_futurework}

This paper presents a comprehensive analysis and comparison of the existing models for aligning embedding spaces w.r.t. the research questions mentioned in Sect.~\ref{sec:problemformulation}. This section summarizes the lessons learnt from the comparison and analysis of different algorithms. In order to encode the structural differences of the input sources in entity - word vector spaces, most of the methods exploit the structure of the links between the entities from Wikipedia and propose different ways of capturing the contextual information of a certain entity mention in the text. Also, some models are trained with pre-aligned triples and sentences. However, the relations between the entities which are encoded in the KGs remain unexplored for most of the methods. On the other hand, models for vector space alignment of entities across multiple KGs follow supervised learning mechanisms and require a set of pre-aligned entities or triples as initial seeds. Not all the information encoded in the KGs is exploited in these models, such as multi-valued object and attribute relations. This also points to RQ3 but the methods do not consider the equivalent relations across KGs.
As future work, these models could be further exploited to build various NLP applications such as machine translation. Furthermore, the challenges mentioned in the paper paved the way for promising research directions in alignment of embedding spaces.

%
%
%
\bibliographystyle{splncs04}
\bibliography{bibliographies.bib}

\begin{thebibliography}{10}
\providecommand{\url}[1]{\texttt{#1}}
\providecommand{\urlprefix}{URL }
\providecommand{\doi}[1]{https://doi.org/#1}

\bibitem{DBLP:conf/emnlp/ArtetxeLA16}
Artetxe, M., Labaka, G., Agirre, E.: Learning {P}rincipled {B}ilingual
  {M}appings of {W}ord {E}mbeddings while {P}reserving {M}onolingual
  {I}nvariance. In: Proceedings of Conference on Empirical Methods in Natural
  Language Processing (2016)

\bibitem{bengio2013representation}
Bengio, Y., Courville, A., Vincent, P.: Representation learning:{A} {R}eview
  and {N}ew {P}erspectives. IEEE transactions on pattern analysis and machine
  intelligence  (2013)

\bibitem{bollacker2008freebase}
Bollacker, K., Evans, C., Paritosh, P., Sturge, T., Taylor, J.: Freebase: a
  collaboratively created graph database for structuring human knowledge. In:
  Proceedings of the 2008 ACM SIGMOD international conference on Management of
  data (2008)

\bibitem{DBLP:conf/nips/BordesUGWY13}
Bordes, A., Usunier, N., Garc{\'{\i}}a{-}Dur{\'{a}}n, A., Weston, J.,
  Yakhnenko, O.: Translating {E}mbeddings for {M}odeling {M}ulti-relational
  {D}ata. In: 27th Annual Conference on Neural Information Processing Systems
  (2013)

\bibitem{DBLP:conf/coling/CaoZZM16}
Cao, H., Zhao, T., Zhang, S., Meng, Y.: A {D}istribution-based {M}odel to learn
  {B}ilingual {W}ord {E}mbeddings. In: 26th International Conference on
  Computational Linguistics (2016)

\bibitem{cao2018joint}
Cao, Y., Hou, L., Li, J., Liu, Z., Li, C., Chen, X., Dong, T.: Joint
  {R}epresentation {L}earning of {C}ross-lingual {W}ords and {E}ntities via
  {A}ttentive {D}istant {S}upervision. In: In Proceddings of the Empirical
  Methods in Natural Language Processing (2018)

\bibitem{cao2017bridge}
Cao, Y., Huang, L., Ji, H., Chen, X., Li, J.: Bridge {T}ext and {K}nowledge by
  {L}earning {M}ulti-prototype {E}ntity {M}ention {E}mbedding. In: Proceedings
  of the 55th Annual Meeting of the Association for Computational Linguistics
  (2017)

\bibitem{chen2017multilingual}
Chen, M., Tian, Y., Yang, M., Zaniolo, C.: Multilingual {K}nowledge {G}raph
  {E}mbeddings for {C}ross-lingual {K}nowledge {A}lignment. In: Proceedings of
  the 26th International Joint Conference on Artificial Intelligence (2017)

\bibitem{conneau2017word}
Conneau, A., Lample, G., Ranzato, M., Denoyer, L., J{\'e}gou, H.: Word
  translation without parallel data. arXiv preprint arXiv:1710.04087  (2017)

\bibitem{DBLP:conf/eacl/FaruquiD14}
Faruqui, M., Dyer, C.: Improving {V}ector {S}pace {W}ord {R}epresentations
  using {M}ultilingual {C}orrelation. In: Proceedings of the 14th Conference of
  the European Chapter of the Association for Computational Linguistics, {EACL}
  (2014)

\bibitem{gabrilovich2013facc1}
Gabrilovich, E., Ringgaard, M., Subramanya, A.: Facc1: Freebase annotation of
  clueweb corpora. Version  (2013)

\bibitem{han2016joint}
Han, X., Liu, Z., Sun, M.: Joint representation learning of text and knowledge
  for knowledge graph completion. arXiv preprint arXiv:1611.04125  (2016)

\bibitem{hao2016joint}
Hao, Y., Zhang, Y., He, S., Liu, K., Zhao, J.: A {J}oint {E}mbedding method for
  {E}ntity {A}lignment of {K}nowledge {B}ases. In: China Conference on
  Knowledge Graph and Semantic Computing (2016)

\bibitem{DBLP:conf/aaai/LinLSLZ15}
Lin, Y., Liu, Z., Sun, M., Liu, Y., Zhu, X.: Learning {E}ntity and {R}elation
  {E}mbeddings for {K}nowledge {G}raph {C}ompletion. In: Proceedings of the
  Twenty-Ninth {AAAI} Conference on Artificial Intelligence (2015)

\bibitem{DBLP:journals/corr/abs-1301-3781}
Mikolov, T., Chen, K., Corrado, G., Dean, J.: Efficient {E}stimation of {W}ord
  {R}epresentations in {V}ector {S}pace. In: 1st International Conference on
  Learning Representations, {ICLR}, Workshop Track Proceedings (2013)

\bibitem{DBLP:journals/corr/MikolovLS13}
Mikolov, T., Le, Q.V., Sutskever, I.: Exploiting {S}imilarities among
  {L}anguages for {M}achine {T}ranslation. CoRR  (2013)

\bibitem{DBLP:conf/esws/MorenoBBDLRTG17}
Moreno, J.G., Besan{\c{c}}on, R., Beaumont, R., D'hondt, E., Ligozat, A.,
  Rosset, S., Tannier, X., Grau, B.: Combining {W}ord and {E}ntity {E}mbeddings
  for {E}ntity {L}inking. In: The Semantic Web - 14th International Conference,
  {ESWC} (2017)

\bibitem{sun2017cross}
Sun, Z., Hu, W., Li, C.: Cross-lingual {E}ntity {A}lignment via {J}oint
  {A}ttribute-{P}reserving {E}mbedding. In: International Semantic Web
  Conference (2017)

\bibitem{sun2018bootstrapping}
Sun, Z., Hu, W., Zhang, Q., Qu, Y.: Bootstrapping {E}ntity {A}lignment with
  {K}nowledge {G}raph {E}mbedding. In: IJCAI (2018)

\bibitem{toutanova2015representing}
Toutanova, K., Chen, D., Pantel, P., Poon, H., Choudhury, P., Gamon, M.:
  Representing text for joint embedding of text and knowledge bases. In:
  Proceedings of the 2015 Conference on Empirical Methods in Natural Language
  Processing (2015)

\bibitem{trisedya2019entity}
Trisedya, B.D., Qi, J., Zhang, R.: Entity alignment between knowledge graphs
  using attribute embeddings. In: Proceedings of the AAAI Conference on
  Artificial Intelligence (2019)

\bibitem{wang2018cross}
Wang, Z., Lv, Q., Lan, X., Zhang, Y.: Cross-lingual knowledge graph alignment
  via graph convolutional networks. In: Proceedings of the 2018 Conference on
  Empirical Methods in Natural Language Processing (2018)

\bibitem{DBLP:conf/naacl/XingWLL15}
Xing, C., Wang, D., Liu, C., Lin, Y.: Normalized {W}ord {E}mbedding and
  {O}rthogonal {T}ransform for {B}ilingual {W}ord {T}ranslation. In: North
  American Chapter of the Association for Computational Linguistics: Human
  Language Technologies (2015)

\bibitem{yamada2016joint}
Yamada, I., Shindo, H., Takeda, H., Takefuji, Y.: Joint {L}earning of the
  {E}mbedding of {W}ords and {E}ntities for {N}amed {E}ntity {D}isambiguation.
  CoNLL 2016  (2016)

\bibitem{yang2014embedding}
Yang, B., Yih, W.t., He, X., Gao, J., Deng, L.: Embedding entities and
  relations for learning and inference in knowledge bases. arXiv preprint
  arXiv:1412.6575  (2014)

\bibitem{zhu2017iterative}
Zhu, H., Xie, R., Liu, Z., Sun, M.: Iterative {E}ntity {A}lignment via {J}oint
  {K}nowledge {E}mbeddings. In: IJCAI (2017)

\end{thebibliography}

\end{document}